\documentclass{ieeetj}

\usepackage{amsmath,amssymb,amsfonts}
\usepackage{algorithmic}
\usepackage{graphicx,color}
\usepackage{multirow}
\usepackage{subcaption}
\usepackage[numbers,sort&compress]{natbib}
\usepackage{textcomp}
\usepackage{xcolor}
\usepackage{hyperref}
\hypersetup{hidelinks}
\usepackage{algorithm,algorithmic}
\def\BibTeX{{\rm B\kern-.05em{\sc i\kern-.025em b}\kern-.08em
    T\kern-.1667em\lower.7ex\hbox{E}\kern-.125emX}}
\AtBeginDocument{\definecolor{tmlcncolor}{cmyk}{0.93,0.59,0.15,0.02}\definecolor{NavyBlue}{RGB}{0,86,125}}

\def\OJlogo{\vspace{-4pt}$<$Society logo(s) and publication title will appear here.$>$}
\def\seclogo{\vspace{10pt}$<$Society logo(s) and publication title will appear here.$>$}

\def\authorrefmark#1{\ensuremath{^{\textbf{#1}}}}

\begin{document}
\receiveddate{XX Month, XXXX}
\reviseddate{XX Month, XXXX}
\accepteddate{XX Month, XXXX}
\publisheddate{XX Month, XXXX}
\currentdate{XX Month, XXXX}
\doiinfo{XXXX.2022.1234567}

\markboth{}{Author {et al.}}

\title{Motor Imagery Classification Using Feature Fusion of Spatially Weighted Electroencephalography}



\author{Abdullah Al Shiam\authorrefmark{1}, Md. Khademul Islam Molla\authorrefmark{2},\\Abu Saleh Musa Miah\authorrefmark{2,3} and  Md. Abdus Samad Kamal\authorrefmark{4}}
\affil{Department of Computer Science and Engineering, Netrokona University, Bangladesh (e-mail: shiam.cse@neu.ac.bd)}
\affil{Department of Computer Science and Engineering, University of Rajshahi, Bangladesh (e-mail: khademul.cse@ru.ac.bd)}
\affil{Graduate Dept. of Computer Science and Engineering, University of Aizu, Japan (e-mail: musa@u-aizu.ac.jp)}
\affil{Cluster of Electronics and Mechanical Engineering, Graduate School of Science and Technology, Gunma University, Kiryu, Gunma 376-8515, Japan (e-mail: maskamal@gunma-u.ac.jp)}
\corresp{Corresponding author: Md. Abdus Samad Kamal (email: maskamal@gunma-u.ac.jp), Md. Khademul Islam Molla (khademul.cse@ru.ac.bd).}
\authornote{This work is a part of a project supported by Netrokona University, Netrokona 2400, Bangladesh.}

\begin{abstract}
A Brain-Computer Interface (BCI) connects the human brain to the outside world, providing a direct communication channel. Electroencephalography (EEG) signals are commonly used in BCIs to reflect cognitive patterns related to motor function activities. However, due to the multichannel nature of EEG signals, explicit information processing is crucial to lessen computational complexity in BCI systems. This study proposes an innovative method based on brain region-specific channel selection and multi-domain feature fusion to improve classification accuracy.
The novelty of the proposed approach lies in region-based channel selection, where EEG channels are grouped according to their functional relevance to distinct brain regions. By selecting channels based on specific regions involved in motor imagery (MI) tasks, this technique eliminates irrelevant channels, reducing data dimensionality and improving computational efficiency. This also ensures that the extracted features are more reflective of the brain's actual activity related to motor tasks. Three distinct feature extraction methods—Common Spatial Pattern (CSP), Fuzzy C-means clustering, and Tangent Space Mapping (TSM)—are applied to each group of channels based on their brain region. Each method targets different characteristics of the EEG signal: CSP focuses on spatial patterns, Fuzzy C-means identifies clusters within the data, and TSM captures non-linear patterns in the signal. The features extracted from these three methods are complementary, allowing for a richer representation of the EEG data. These individual features are then integrated into a high-dimensional feature vector, enhancing the robustness of the model.
The combined feature vector is used to classify motor imagery tasks (left hand, right hand, and right foot) using Support Vector Machine (SVM). The proposed method was validated on publicly available benchmark EEG datasets (IVA and I) from the BCI competition III and IV. The results show that the approach outperforms existing methods, achieving classification accuracies of 90.77\% and 84.50\% for datasets IVA and I, respectively. This study demonstrates that combining region-based channel selection with multi-domain feature extraction significantly enhances the performance and interpretability of EEG-based BCI systems.
\end{abstract}



\maketitle

\section{INTRODUCTION}
\label{sec:introduction}
\IEEEPARstart{I}{n} order to control peripheral tasks like a cursor or an artificial limb, a Brain-Computer Interface (BCI) establishes a connection between the brain and an outside device \cite{b1,kabir2023investigating_miah_SVM_LDA_MLP,miah2019eeg}. The brain and the control gadget are able to communicate directly through this interface. Instead of using the conventional path—the neuromuscular system—to guide the mouse's finger, the control signals are sent straight from the brain to the mechanism controlling the cursor. It does this by reading neural impulses sent by the human brain and translating them into physical commands using software and computer hardware. For paralyzed patients, this enables controlling devices, motorized wheelchairs, or prosthetics solely through thought \cite{b1,kabir2023investigating_miah_SVM_LDA_MLP,miah2019eeg}. While current BCIs require conscious effort, future applications, like prosthetic control, aim for seamless operation. Minimally invasive electrode devices and surgical techniques are vital for advancing BCI technology. The brain assimilates and controls implanted mechanical devices as if they were body extensions. Ongoing research is exploring non-invasive BCIs. The role of BCIs involves processing incoming sensory stimuli through peripheral nerves, triggering actions by transmitting impulses to actuators (muscles or glands). BCIs hold potential for both humans and animals, with increasing interest from healthy individuals for rehabilitation and hands-free gaming. However, accessibility remains a challenge. BCIs, though seemingly new, were first proposed by Jacques J. Vidal in 1973 \cite{b4}. The BCI system processes brain signals and turns them into actions, demonstrated by a person manipulating a pinball machine using only their brain, with Electroencephalography (EEG) capturing brain activity \cite{b5}. The EEG cap records electrical activity, which is analyzed by a computer to produce actions \cite{miah2022movie_miah_SVM,miah2021event_EEG}.

The computer can translate a subject's imagined left or right hand movements into corresponding signals whenever those movements are imagined. The EEG system measures brain activity by acquiring signals \cite{miah2022natural_EEG_SVM_KNN}. This process involves collecting cerebral signals and converting them into data for computational analysis. The second phase processes the EEG data into output signals, while the third step extracts relevant features that describe the signals effectively \cite{miah2017motor_miah_SVM_PCA_ANOVA}. Once the features are extracted, classification occurs \cite{kabir2023investigating_miah_SVM_LDA_MLP,kabir2024exploring_miah_LDA}. Feedback can take various forms, such as a pinball flipper, prosthesis, virtual keyboard, or steering wheel \citep{b7, b8, b9}. The key feature of feedback is its ability to convert the output signal into a designated action. BCI experts focus on understanding the brain's response to specific stimuli, with the BCI system acting as a high-tech measuring device, not a tool for directly converting thoughts into actions. An experiment can demonstrate memory effectiveness. While wearing an EEG cap, a subject is shown Chinese characters, aiming to maximize memorization effectiveness while the EEG records brain activity. After memorization, the participant evaluates their memory through a pen-and-paper test, bypassing the EEG. The researcher attempts to find patterns in EEG signals that correspond to correctly and incorrectly memorized symbols. By predicting the learning process's efficacy, insights into the stimulus can be gained. A BCI system cannot read minds or understand mental processes. It is limited to classifying certain patterns of brain activity, with attention and motor imagery being the most effective patterns for BCI applications.

 Despite significant advancements in Brain-Computer Interfaces (BCIs), the accurate and robust decoding of user intent from EEG signals remains a major challenge because of the complex spatial-temporal dynamics, non-stationarity, and reduced signal-to-noise ratio of brain activity \cite{zobaed2020real}.  Most current BCI systems depend on features derived from singular domains (e.g., temporal, spectral, or spatial), often neglecting the complementary information available across multiple domains. Furthermore, traditional approaches typically treat the brain as a homogeneous signal source, ignoring functional specialization across brain regions. This leads to suboptimal feature representations and limited classification performance. Therefore, there is a critical need for an integrated approach that combines multi-domain features with brain region-specific knowledge to enhance the interpretability and efficacy of EEG based brain-computer interfaces.

This project develops a multi-domain feature integration framework for EEG-based Brain-Computer Interfaces (BCIs), grounded in brain area analysis. The proposed approach extracts and combines discriminative features from spatial, temporal, and spectral domains, based on the functional relevance of distinct brain regions. By integrating multiple feature types and incorporating brain-region specificity, the goal is to enhance the classification accuracy, robustness, and interpretability of BCI systems. Key contributions of this research include:

\begin{itemize}
\item \textbf{Brain Region-Based Categorization:} A method for categorizing EEG channels according to their corresponding brain regions, improving the efficiency of multi-channel EEG signal processing.
\item \textbf{Multi-Domain Feature Fusion:} The use of three feature extraction techniques—Common Spatial Pattern (CSP), Fuzzy C-Means clustering, and Tangent Space Mapping (TSM)—to extract robust features from each channel group, capturing complex brain activity patterns.
\item \textbf{Feature Integration for Classification:} Fusion of features into a high-dimensional vector used Support Vector Machine (SVM) classification for distinguishing motor imagery tasks (left/right hand, right foot) using.
\item \textbf{Experimental Validation:} Validation using publicly available benchmark datasets from the BCI competition III and IV, demonstrating superior classification performance with accuracies of 90.77\% and 84.50\% for datasets IVA and I, respectively.
\end{itemize}

\section{Literature Review}
Motor imagery (MI), where users imagine limb or muscle movements without physical execution, is crucial in EEG-based brain-computer interface (BCI) systems. MI induces oscillatory patterns in sensorimotor rhythms, detected as event-related synchronization (ERS) and event-related desynchronization (ERD) \cite{b10, b11}. To improve classification, BCI systems typically record EEG signals from multiple electrodes across the scalp \cite{b12, b13}. Recent studies have shifted from isolated electrodes to structured brain activity representations. This study proposes a brain-region-based multi-domain feature fusion strategy for binary classification of MI EEG signals. The approach groups electrodes by brain regions (e.g., left and right motor cortex) and extracts statistical, spectral, and spatial features from each region, which are then fused to create a comprehensive, physiologically informed representation of MI activity. While prior MI-BCI studies focused on motor cortex activity using datasets from BCI Competitions III and IV, this method leverages multi-domain features from spatially meaningful brain regions to improve classification performance and interpretability. Previous work, such as BECSP \cite{b14} and adaptive LASSO \cite{b15}, introduced novel feature extraction and selection techniques. Additionally, WOLA filtering and linear discriminant analysis \cite{b16} have been used for signal classification. This study enhances motor imagery EEG classification through multi-scale CNN fusion, using spatial and temporal filters to improve performance on the BCI IV 2a dataset compared to existing methods \cite{b17}.
The spatio-spectral feature representation method in \cite{b18} employs 3-D feature maps and a layer-wise decomposition model within a 3-D CNN framework, achieving average accuracies of 87.15\%, 75.85\%, and 70.37\% across various BCI datasets. This method preserves the EEG feature space's multivariate structure and interdependencies. The 3D-CNN approach, with its layer-wise decomposition, produces accurate classification results using only one trial. Li et al. \cite{b19} introduced the P-3DCNN, which achieves an average decoding accuracy of 86.69\% for motor imagery tasks. This network uses cubic interpolation along with 1D and 2D convolutional layers to generate spatial-frequency maps from EEG data. Moein et al. \cite{b20} proposed a feature fusion approach combining spatial, spectral, and temporal domains. Using Constant-Q filters, they improved temporal resolution for both lower and higher-frequency subbands, and feature selection algorithms identified the best features from the combined set. Hongli et al. \cite{b21} introduced a hybrid CNN-LSTM model for feature fusion, where convolutional and LSTM layers jointly extract spatial and temporal features, which are then combined into a feature vector for classification. Singh et al. \cite{b22} proposed using spatially regularized symmetric positive-definite (SPD) matrices to reduce motor imagery calibration time. Their system mitigates high dimensionality by using spatial filters regularized with previous EEG data to reduce SPD matrix dimensions. Park et al. \cite{b23} developed a filter bank-based feature extraction technique to address CSP's reliance on sample-based covariance. Their approach segments EEG data, applies regularized CSP (R-CSP), and selects features using mutual information, followed by ensemble classification based on detected features.
Xinzhi et al. \cite{b24} introduced a novel deep learning network that integrates multi-modal temporal information and global dependencies. This network combines CNNs with a self-attention strategy. Initially, the network extracts multi-modal temporal features from two perspectives: average and variance. A self-attention module then captures global interdependencies between these two feature dimensions. A convolutional encoder merges the average-pooled and variance-pooled data to form more discriminative features. To improve generalization, the network incorporates a signal segmentation and recombination data augmentation technique.
Singh et al. \cite{b25} focused on reducing calibration time using a Riemannian method. This method addresses performance degradation when sample sizes are small due to the high dimensionality of covariance matrices relative to trial numbers. Their approach reduces dimensionality by regularizing spatial filters using data from additional participants. Zhang et al. \cite{b26} presented a technique for classifying motor imagery EEG signals using multi-kernel ELM (MKELM). By combining polynomial and Gaussian kernels in a multi-kernel learning strategy, MKELM extracts additional information from nonlinear feature spaces, enhancing EEG classification accuracy. Various methods, including discriminative filter bank CSP, sparse filter-bank CSP, subband CSP, and filterbank CSP, have been used for classifying narrowband EEG signals in motor imagery tasks. This work employs a multi-domain feature fusion method based on brain regions to extract meaningful features for robust classification of motor imagery tasks.
EEG channels are grouped based on the left and right hemispheres, corresponding to motor responses. Two EEG channel groups are created, and three feature extraction methods—CSP, Fuzzy C-means, and TSM—are applied to extract features from each group. These features are then combined into a single feature vector, resulting in a higher-dimensional feature space \cite{b32}. However, some features may lack accuracy, distorting the machine learning algorithm and reducing performance \cite{b33}. To improve classification reliability, a subset of the most relevant features must be selected, eliminating extraneous ones to reduce overfitting, enhance performance, and decrease training time. This study aims to improve motor imagery (MI) classification performance by combining features and applying embedded feature selection using the Random Forest (RF) approach to identify the most discriminative features. The selected feature vector is then used for MI classification tasks with SVM, MLP, RF, and XGBoost classifiers. The methodology was tested using the BCI Competition III IVA and BCI Competition IV I datasets. Compared to current MI classification techniques, the proposed method yields better classification accuracy.

\section{DATASETS}
In the study, we have used two EEG-based MI datasets, which are described below: 
\subsubsection{BCI competition III Dataset IVA}\label{BCI-dataset-IVA}
\begin{figure}[t!]
\centering
\includegraphics[width=0.35\textwidth]{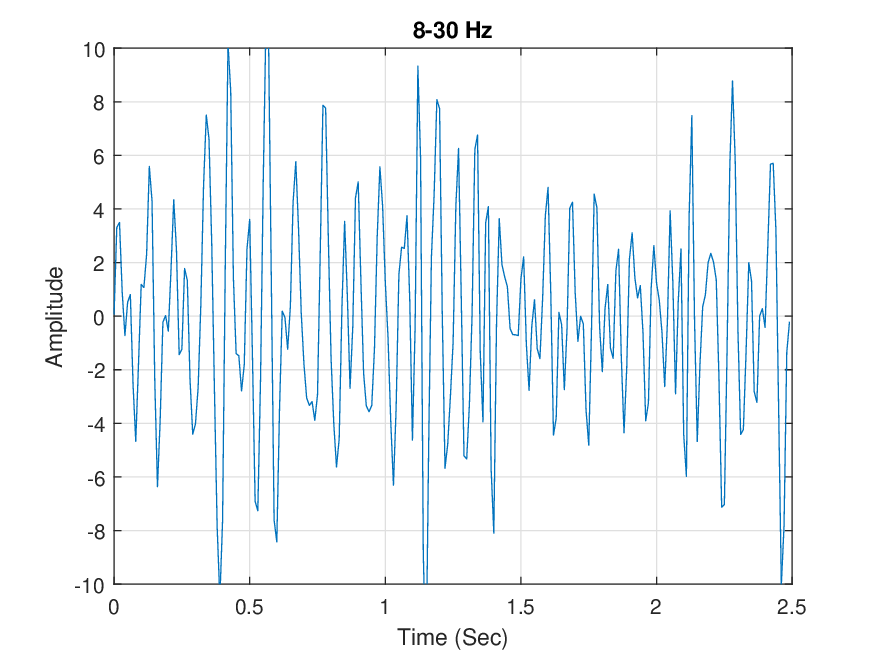}
\caption{Electroencephalography (EEG) of subject aw from channel C1 of BCI competition III IVA Dataset.}
\label{fig:EEG_BcI_III_Iva}
\end{figure}

Our experiment utilizes the publicly available motor imagery (MI) dataset IVA from the BCI competition III, detailed in \cite{b34}. The dataset includes EEG signals from five healthy individuals (aa, al, aw, av, ay) who participated in motor imagery task trials. Participants, aged 24-25, sat comfortably in a chair with their arms on the armrests during the recording. They were instructed to keep their arms and hands relaxed and not move their eyes during recordings. Each participant completed 280 runs, divided into four sessions of 70 runs. Visual stimuli were presented while participants imagined moving their right hand or foot for 3.5 seconds. Goal signals were given between 1.75 to 2.25 seconds, allowing participants to relax while focusing on one of the two tasks.
EEG signals were recorded at 118 worldwide 10/20 locations, using a band-pass filter from 0.05 to 200 Hz and digitized at 1000 Hz with 16-bit resolution. The dataset was then downsized to 100 Hz for processing. Figure \ref{fig:EEG_BcI_III_Iva} shows the single-channel dataset visualization, and Figure \ref{fig_timing_BCI_III} illustrates the timing diagram of the BCI competition III IVA Dataset.

\begin{figure}[t!]
\centering
\includegraphics[width=0.35\textwidth]{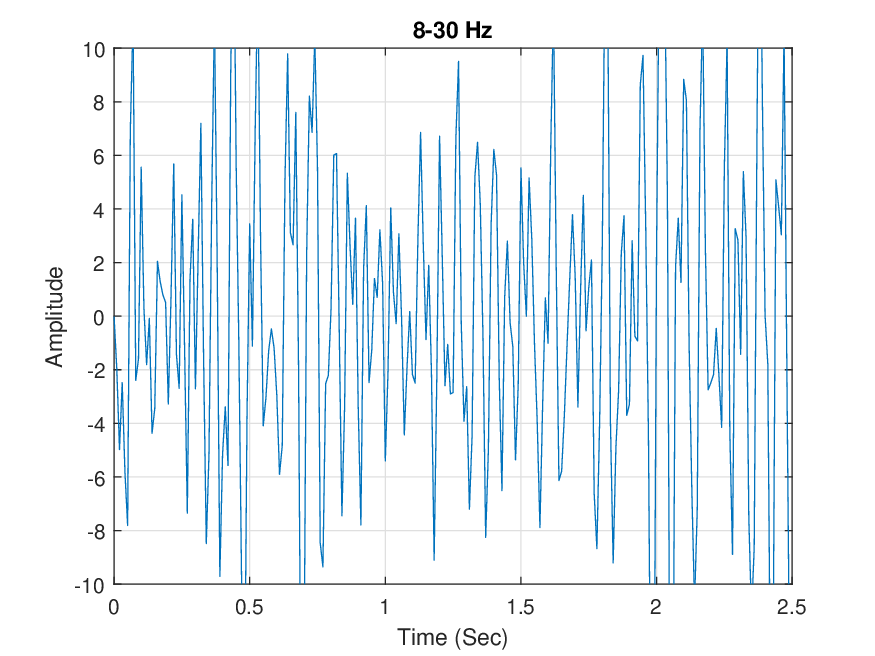}
\caption{Electroencephalography (EEG) of subject a from channel C1 of BCI competition IV-1 Dataset.}
\label{fig:EEG_BCI_IV_I}
\end{figure}
Experimental data comes from electroencephalograms recorded between half a second and three seconds after the cue.  We identify the first half-second (0-0.5 s) and the last half-second (3-3.5 s) as the pre- and post-imagination periods, respectively. 
\begin{figure}[t!]
\centering
\includegraphics[width=0.35\textwidth]{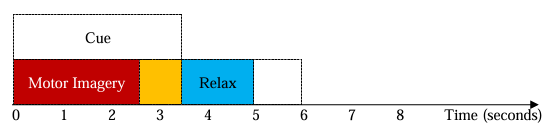}
\caption{BCI Competition III-IVA dataset timing sequence.}
\label{fig_timing_BCI_III}
\end{figure}

\subsubsection{BCI competition IV Dataset I}
\begin{figure}[t!]
\centering
\includegraphics[width=0.35\textwidth]{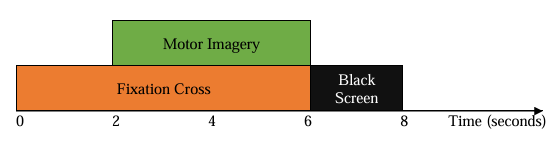}
\caption{BCI competition IV I dataset timing sequence.}
\label{timing_IV_I}
\end{figure}

We also utilized another benchmark dataset, namely the BCI competition IV Dataset 1\cite{b43}, which consisted of the 7 healthy people, namely a, b, c, d, e, f, and g, who participated in the MI task. For this dataset, a 100 Hz sampling frequency is used. Two out of the three categories of motor imagery tasks, left-hand, right-hand, and foot movement, have been executed by each individual. The person completed a particular mental exercise while four seconds of arrow-shaped visual signals were shown on the screen.  There was a two-second fixation cross and two seconds of blank screen, interspersed with the signals.   The fixation cross was shown over the cues for a duration of six seconds. Consequently, there is a 4-second rest period after each trial, which lasts 4 seconds. Each subject has a total of 200 trials, with 100 trials for each of the two classes selected for that subject.  400 samples are included in a trial that lasts 4 seconds (4 seconds x 100 Hz).  Figure \ref{fig:EEG_BCI_IV_I} visualizes the single data information of this dataset. Figure \ref{timing_IV_I} shows the data paradigm for the BCI Competition IV I dataset.

\section{METHODOLOGY}
\label{sec:guidelines}
\begin{figure*}[t!]
    \centering
    \includegraphics[width=0.8\textwidth]{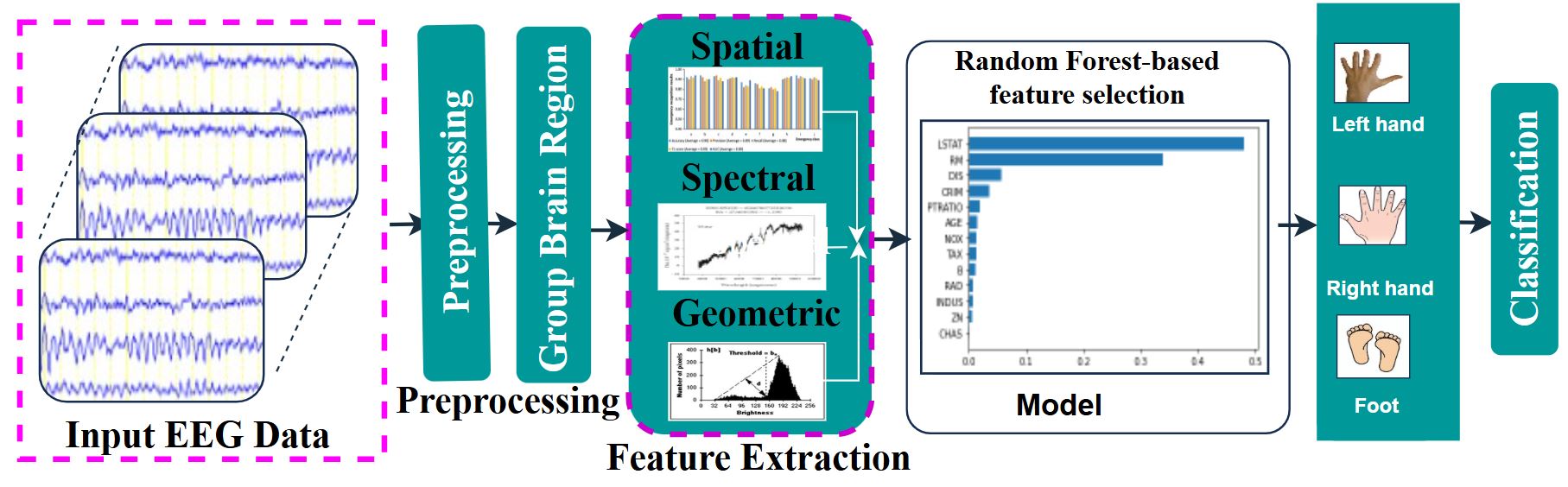}
    \caption{Block diagram of the proposed architecture}
    \label{fig:main_block_diagram}
\end{figure*}

The proposed methodology utilizes brain region-based multi-domain feature integration for classifying EEG signals. The process involves six main steps, which are shown in the Figure \ref{fig:main_block_diagram} and are also described below:
\begin{enumerate}
    \item \textbf{Grouping EEG Channels:} Channels are grouped based on brain regions associated with motor imagery (MI) tasks.
    \item \textbf{Band-Pass Filtering:} Relevant frequency bands are isolated for each brain region group.
    \item \textbf{Feature Extraction:} Spatial, spectral, and geometric features are extracted from each group.
    \item \textbf{Feature Integration:} All extracted features are combined into a comprehensive feature vector for classification.
    \item \textbf{Feature Selection:} Discriminative features are selected using Random Forest-based feature importance.
    \item \textbf{Classification:} Various classifiers, including SVM, MLP, RF, and XGBoost, are trained and evaluated using the selected features from the labeled training dataset.
\end{enumerate}

This approach aims to enhance classification accuracy by leveraging multi-domain features and brain-region specificity. The details of each section are given below. 

 






\subsection{PRE-PROCESSING}
Electrophysiological noise frequently taints the multichannel raw EEG. Furthermore, several narrowband components of the EEG signal show an increased response to the specific MI task \cite{b27, b29}. Frequencies between 8 and 30 Hz contain the most brain activity during MI tasks \cite{b35}. Based on the research in the study, we employed a band-pass filter with a frequency range between 8 and 30 Hz.

\subsection{SPATIALLY WEIGHTED EEG}
EEG recordings typically contain a large number of electrodes distributed across the entire scalp (59 or 118 channels, depending on the dataset used in this study). Not all EEG channels provide the same level of useful information for motor imagery (MI) classification. As MI tasks primarily engage the motor cortex—specifically the primary, supplementary, and premotor cortical regions \cite{b35}—electrodes placed outside these areas often introduce noise and unnecessary computational burden.

\begin{figure}[t!]
    \centering
    \includegraphics[width=0.35\textwidth]{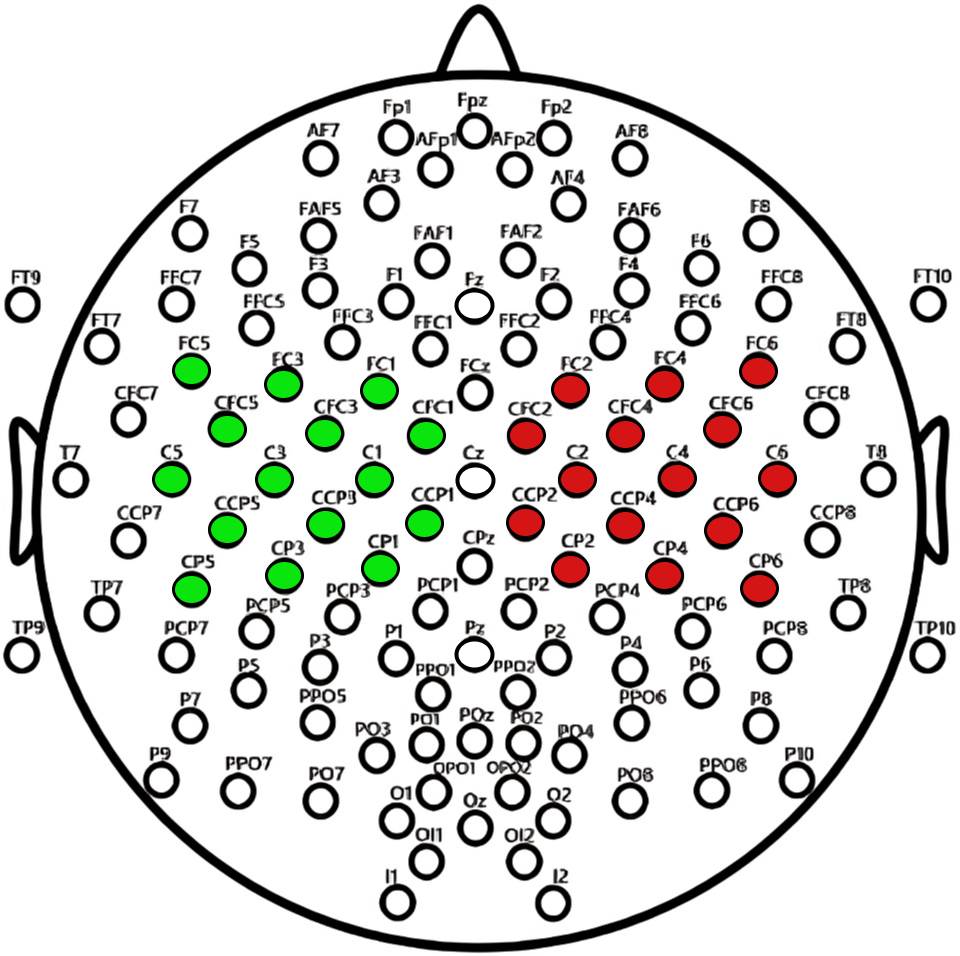}
    \caption{A selection of thirty electrodes, highlighted in green (left hemisphere) and red (right hemisphere), has been chosen from a total of 118 for the purpose of experimental evaluation in this study.}
    \label{fig:brain_regision_electrodes}
\end{figure}

\begin{figure*}[t!]
    \centering
    \begin{subfigure}{0.24\textwidth}
        \centering
        \includegraphics[width=\linewidth, height=2.5cm]{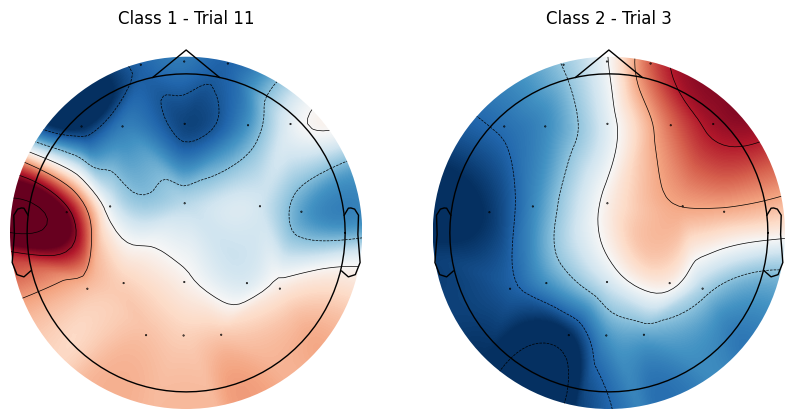}
        \caption{Subject ay}
    \end{subfigure}\hfill
    \begin{subfigure}{0.24\textwidth}
        \centering
        \includegraphics[width=\linewidth, height=2.5cm]{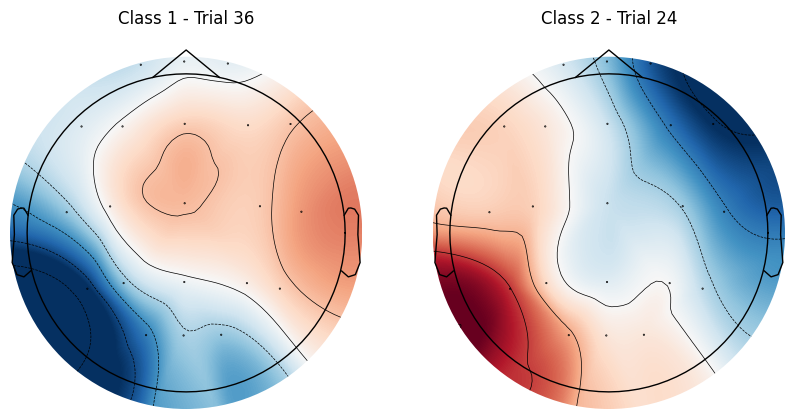}
        \caption{Subject al}
    \end{subfigure}\hfill
    \begin{subfigure}{0.24\textwidth}
        \centering
        \includegraphics[width=\linewidth, height=2.5cm]{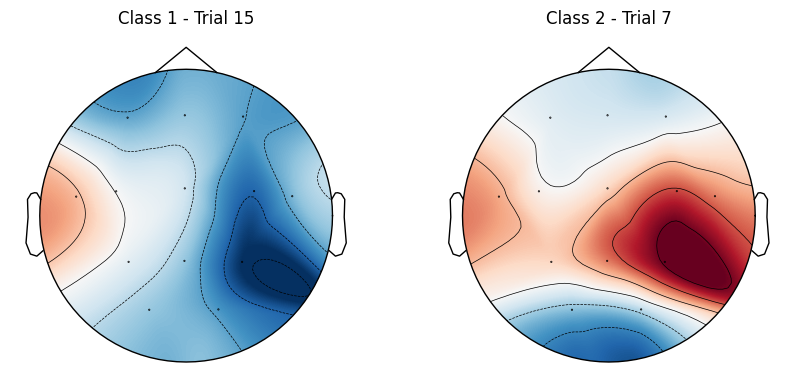}
        \caption{Subject a}
    \end{subfigure}\hfill
    \begin{subfigure}{0.24\textwidth}
        \centering
        \includegraphics[width=\linewidth, height=2.5cm]{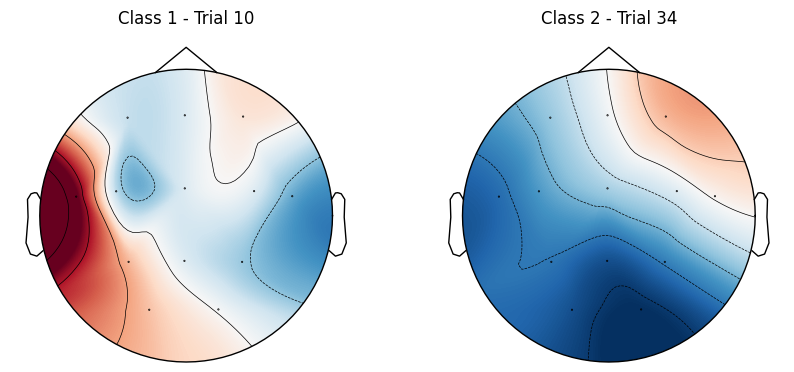}
        \caption{Subject b}
    \end{subfigure}
    \caption{Topographic distribution of EEG potentials for subjects ay, al, a, and b from both datasets during motor imagery tasks. A clear lateralized activation is observed, with stronger cortical modulation over the left sensorimotor cortex, corresponding to right-hand motor imagery, while central regions reflect lower activity related to foot imagery.}
    \label{fig7}
\end{figure*}

To address this, we introduce a spatial weighting strategy that emphasizes task-relevant regions while suppressing irrelevant signals. In the proposed method, each EEG channel is assigned a binary spatial weight, defined as \( w_i \in \{0,1\} \) for channel \( i \). A weight of \( w_i = 1 \) indicates that the channel is functionally relevant and lies over the sensorimotor cortex, while \( w_i = 0 \) excludes channels from further processing. Formally, the weighted EEG signal can be expressed as  
\[
X_w = W \odot X,
\]
where \( X \) represents the original EEG signal matrix, \( W = \text{diag}(w_1, w_2, \dots, w_N) \) is the binary weighting matrix, and \( \odot \) denotes element-wise multiplication. This process reduces the full electrode set to approximately 30 channels concentrated over the left and right motor cortices (Figure~\ref{fig:brain_regision_electrodes}). Grouping the weighted channels according to their anatomical location allows the construction of localized representations of brain activity, ensuring that the retained signals correspond to MI-related neural patterns.  
Figure~\ref{fig7} illustrates topomaps of single-trial EEG activity, highlighting the effectiveness of this approach. For instance, Class 2 (right foot) produces stronger activation in the left motor cortex, while Class 1 (left hand/right hand) elicits enhanced activity in the right motor cortex. Although trial-to-trial variability is present in both amplitude and spatial distribution, the weighted channel selection consistently reveals physiologically meaningful activation patterns. By eliminating irrelevant channels and emphasizing localized activity, the spatial weighting process not only improves classification accuracy but also reduces computational complexity in EEG-based BCI systems.  

\subsection{FEATURE EXTRACTION}
A feature is a unique property or attribute of an object that can be measured. Extracting descriptive, discriminating, and independent features is a critical step in developing successful machine learning pattern classification algorithms. The feature-extraction approach is useful for reducing the amount of data that needs to be processed without losing relevant information. For a given analysis, it reduces the amount of redundant data.

\subsubsection{Common Spatial Pattern (CSP)}
CSP method decomposes EEG signals using spatial filters to optimize class variance, reduce intra-class variation, and enhance inter-class discrimination for effective EEG classification \cite{joy2020multiclass, b36}. To illustrate the CSP algorithm, we will analyze the original EEG data matrix from the trial corresponding to the target class. Let \( X_h^{\mathrm{c_1}} \) and \( X_f^{\mathrm{c_2}} \) denote the training trials of EEG signals related to hand and foot movements, respectively, using the chosen channels from classes \( c_1 \) and \( c_2 \). The next step is to compute the normalized spatial covariance of the EEG for both hand and foot movements as follows:
\begin{equation}
C_h = \frac{E_h E_h^T}{t_r (E_h E_h^T)}, \quad
C_f = \frac{E_f E_f^T}{t_r (E_f E_f^T)}
\end{equation}

The sum of diagonal elements of a matrix is represented as Tr(.) \cite{b8}.  The CSP method diagonalizes both classes' covariance matrices to extract features.  EEG data are transformed into a projection matrix using a spatial filter $w=R^l$ to maximize variances between classes \cite{b8}.
\begin{equation} \label{eq2}
\mathbf{v} = \arg\max \frac{\mathbf{v}^T \mathbf{S}_1 \mathbf{v}}{\mathbf{v}^T \mathbf{S}_2 \mathbf{v}} 
\quad \text{s.t.} \quad \|\mathbf{v}\|_2 = 1
\end{equation}

The optimal solution to equation \ref{eq2} is attained by addressing an extensive eigenvalue problem.  The greatest and smallest eigenvalues $W$ produced by the eigenvectors are incorporated into the spatial filter $v=\left[v_1, v_2, \ldots, v_{2 W}\right] \in R^{l \times 2 W}$.  The feature vector for any trial is formulated as $f^s=\left[f_1, f_2, \ldots, f_{2 W}\right]$.  The logarithmic transformation is employed to normalize the components of $f_w$
\begin{equation}
f_w=\log \left(\operatorname{var}\left(v_w^T x\right)\right), \quad w=1,2, \ldots, 2 V,
\end{equation}
where var(.) denotes the variance. Using the CSP, we can extract some properties from each subband.   The feature vector for the particular trial is formed by concatenating the features obtained from each subband as $F = [f^1, f^2, \ldots, f^n]$, where $n$ is the number of features.   The MI tasks are classified using a classifier with the help of the feature vector F.

\subsubsection{Fuzzy C Means}
Fuzzy C-Means (FCM) is an unsupervised clustering method that permits each data point to belong to multiple clusters, with varying degrees of membership. Unlike rigid clustering methods like k-means, FCM gives soft labels that better show the differences in EEG signal properties.  This makes FCM a valuable tool for analyzing EEG signals, where the boundaries between mental states are often not clearly defined.  When given a dataset, FCM tries to make the following objective function as small as possible:

\begin{equation}
J_m = \sum_{i=1}^{N} \sum_{j=1}^{K} u_{ij}^m \, \lVert x_i - k_j \rVert^2
\end{equation}
In the context of clustering, \( N \) represents the quantity of data points, while \( K \) denotes the number of clusters. The values 0 and 1 indicate the membership degree of the data item in cluster \( j \), and \( K_j \) refers to the centroid of cluster \( j \). The parameter \( m > 1 \) is the fuzzifier, which controls the degree of fuzziness in the clustering process, with a typical value of \( m = 2 \). The symbol \( || . || \) represents the Euclidean norm. The cluster centers and membership values are updated iteratively using:

\begin{equation}
k_j = \frac{\sum\limits_{i=1}^{N} u_{ij}^m x_i}{\sum\limits_{i=1}^{N} u_{ij}^m}, \quad
u_{ij} = \left( \sum\limits_{k=1}^{K} \left( \frac{ \lVert x_i - k_j \rVert }{ \lVert x_i - k_y \rVert } \right)^{\frac{2}{m-1}} \right)^{-1}
\end{equation}

\subsubsection{Tangent Space Mapping (TSM)}
TSM is a Riemannian geometry-based feature extraction technique, primarily used on Symmetric Positive Definite (SPD) covariance matrices from EEG data.  Because these covariance matrices are on a non-Euclidean manifold, standard vector-based classifiers cannot work directly with them.  To fix this, TSM maps each SPD matrix onto the manifold's tangent space at a reference point, turning the matrix into a Euclidean feature vector.  Let $C^i=R^n*n$ be the covariance matrix for the trial and the Riemannian mean of all covariance matrices.  To find the tangent space projection, do the following:

\begin{equation}
\Delta_i = \bar{C}^{-1/2} . \Big[ \log \big( \bar{C}^{-1/2}  C_i  \bar{C}^{-1/2} \big) \Big] . \bar{C}^{-1/2}
\end{equation}
Here, log represents the matrix logarithm. The resulting matrix is symmetric and can be vectorized (e.g., taking the upper triangular part) to produce a feature vector:

\begin{equation}
x_i = \mathrm{vec}(\Delta_i)
\end{equation}

\subsection{EFFECTIVE FEATURE VECTOR}
Three different domains of features, like spatial, spectral, and geometric features, are computed from the EEG signal. All the feature extracted from the decomposed signal is concatenated to represent a vector form. Therefore, the feature vector for a channel can be deﬁned as:
\begin{equation}
\mathbf{n}_f = [f_1, f_2, \ldots, f_n]
\end{equation}
where n indicates the total number of features (in our case, n=248).

\subsection{DISCREMINATIVE FEATURE SELECTION}
Not every attribute, to a large extent, classification relies on a high-dimensional feature vector.   The performance of the machine learning algorithm could be negatively impacted by specific conditions.   Feature extraction is designed to improve item classification performance by providing useful discriminative information. To identify the optimal set of features from all the available options, feature selection is a crucial step in machine learning. Ridding the feature selection process of superfluous features is another goal while minimizing information loss. This study entails the selection of discriminative features from the raw feature space via embedded feature selection employing the Random Forest Feature Importance approach.  Random Forests generate many decision trees throughout the training phase.  Each tree is constructed using a bootstrapped subset of the data, and at each split, a random selection of characteristics is assessed.  During this procedure, the importance of each feature is evaluated based on its role in reducing impurity.

\subsection{CLASSIFICATION}
We employed Multilayer Perceptron (MLP), XGBoost, Random Forest (RF), and Support Vector Machine (SVM) to evaluate the performance of the proposed method. To assess its effectiveness, 80\% of the data was used for training and 20\% for testing. The classification process utilized both the training and testing datasets, along with the k-fold cross-validation technique, where 
k=5 for each subject.
\section{RESULTS}
In the study, we evaluated the proposed model with two EEG benchmark datasets, namely BCI Competition III and IV datasets, namely IVA and I, respectively.  
\subsection{EXPERIMENTAL SETUP}
Following the aggregation of the MI channel, each dataset sample is segmented into 8-30 Hz signals, from which various features are derived.  A feature vector is generated by amalgamating the attributes from the group.  The extracted features are used to train MLP, SVM, RF, and XGBoost, which are then employed to assess classification performance using the test data. Trials lasting 3 seconds for the BCI competition III IVA Dataset and 4 seconds for the BCI competition IV-1 Dataset are extracted from the EEG data for each participant. The specifics of the data extraction methodology are outlined in the dataset description section. The Common Spatial Pattern (CSP) technique is utilized to extract spatial characteristics from EEG recordings. From each brain region group, one pair of spatial filters is computed, yielding two discriminative features per group. Consequently, for each trial, a 2-dimensional spatial feature vector is obtained from each brain region. In addition to the spatial features, 2-dimensional spectral features and 120-dimensional geometric features are extracted from each region. The spatial, spectral, and geometric features are then concatenated across all brain region groups, resulting in a total of 248 features per trial. These combined features are subsequently used as input to the classifier for decoding the motor imagery tasks. 
Here, we use the k-fold cross-validation method with k set at 5 to evaluate the accuracy of categorization for each subject. For each participant, the dataset is partitioned into k equal portions in a systematic manner. There will be one training group and one testing group out of the (k-1) total. The procedure is carried out k times.   Taking an average of the results from all k iterations yields the classification accuracy. The classification performance is assessed using the formula Acc = 100*(TC/TT), where TT represents the total number of trials in the test dataset, and TC signifies the number of trials accurately identified from TT. Multiple experiments are conducted using the BCI competition III IVA Dataset and the BCI competition IVA-1 Dataset to demonstrate the effectiveness of the proposed approach. To assess the accuracy of the method, several performance metrics, including Precision, Recall, and F1 score, are utilized \cite{miah2017motor_miah_SVM_PCA_ANOVA,miah2019motor_LDA_MAD}.



\begin{figure}[t!]
\centering
\includegraphics[width=0.38\textwidth]{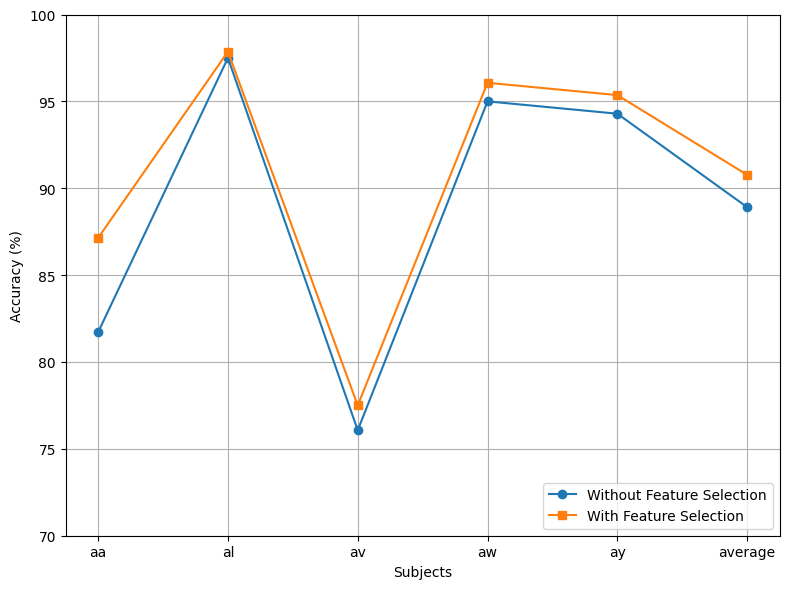}
\caption{Performance (classification accuracy) of individual subjects and their average as a function of with and without feature selection using dataset BCI-Competition III-IVa.} 
\label{feature_selection_vs_without_dataset1}
\end{figure}
\begin{figure}[t!]
\centering
\includegraphics[width=0.38\textwidth]{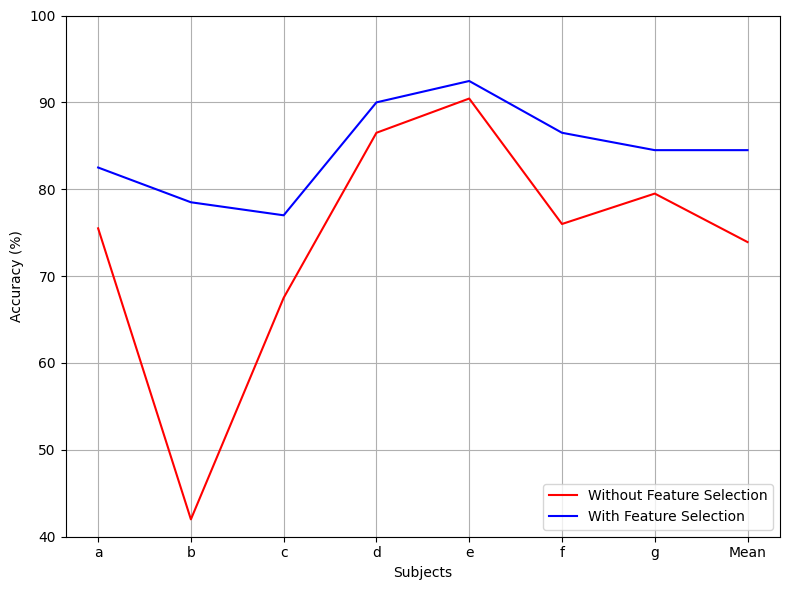}
\caption{Performance (classification accuracy) of individual subjects and their average as a function of with and without feature selection using dataset  BCI-Competition IV-1.} 
\label{feature_selection_vs_without_dataset2}
\end{figure}

As we combined the hybrid feature from 3 different feature extraction modules, we then employed a feature selection module before classification. In the ablation study, we demonstrated the performance with and without feature selection. The number of features selected for MI classification is a crucial determinant of accuracy. Figures \ref{feature_selection_vs_without_dataset1} and \ref{feature_selection_vs_without_dataset2} illustrate the performance of individual subjects with and without feature selection, using the BCI competition III IVA Dataset and the BCI competition IV-1 Dataset. Additionally, the mean values for all subjects are provided. The results show that feature selection consistently yields the highest accuracy across all subjects.

\begin{table}[]
\centering
\caption{The evaluation of the proposed approach in terms of classification accuracy, precision, recall, and F1 score, utilizing MLP, SVM, RF, and XGBoost classifiers on the BCI competition III IVA Dataset, with the most significant outcome highlighted in boldface.}
 \label{tab1} \renewcommand{\arraystretch}{0.7}
\begin{tabular}{|c|c|c|c|c|c|}
\hline
\textbf{Classifier} & \textbf{Subjects} & \textbf{Accuracy} & \textbf{Precision} & \textbf{Recall} & \textbf{F1 Score} \\ \hline
\multirow{5}{*}{MLP} 
& aa & \textbf{91.07} & \textbf{91.00} & \textbf{91.00} & \textbf{91.00} \\ \cline{2-6}
& al & 96.43          & \textbf{97.00} & \textbf{96.00} & \textbf{96.00} \\ \cline{2-6}
& av & 71.43          & 72.00          & 71.00          & 71.00          \\ \cline{2-6}
& aw & \textbf{96.43} & \textbf{97.00} & \textbf{96.00} & \textbf{96.00} \\ \cline{2-6}
& ay & 94.64          & 95.00          & 95.00          & 95.00          \\ \hline
\multicolumn{2}{|c|}{\textbf{Mean}} & 90.00 & \textbf{90.40} & \textbf{89.8} & \textbf{89.8} \\ \hline
\multirow{5}{*}{SVM} 
& aa & 87.14          & \textbf{91.00} & \textbf{91.00} & \textbf{91.00} \\ \cline{2-6}
& al & \textbf{97.86} & \textbf{97.00} & \textbf{96.00} & \textbf{96.00} \\ \cline{2-6}
& av & 77.50          & 71.00          & 70.00          & 70.00          \\ \cline{2-6}
& aw & 96.07          & 95.00          & 95.00          & 95.00          \\ \cline{2-6}
& ay & 95.36          & 97.00          & 96.00          & 96.00          \\ \hline
\multicolumn{2}{|c|}{\textbf{Mean}} & \textbf{90.77} & 90.20 & 89.60 & 89.60 \\ \hline
\multirow{5}{*}{RF} 
& aa & 87.50          & 88.00 & 88.00 & 88.00 \\ \cline{2-6}
& al & 96.43          & \textbf{97.00} & \textbf{96.00} & \textbf{96.00} \\ \cline{2-6}
& av & 73.21          & 74.00 & 74.00 & 73.00 \\ \cline{2-6}
& aw & 91.07          & 92.00 & 91.00 & 91.00 \\ \cline{2-6}
& ay & 92.86          & 94.00 & 93.00 & 93.00 \\ \hline
\multicolumn{2}{|c|}{\textbf{Mean}} & 88.21 & 89.00 & 88.40 & 87.00 \\ \hline
\multirow{5}{*}{XGBoost} 
& aa & 85.71          & 86.00 & 86.00 & 86.00 \\ \cline{2-6}
& al & 96.43          & \textbf{97.00} & \textbf{96.00} & \textbf{96.00} \\ \cline{2-6}
& av & \textbf{78.57} & \textbf{79.00} & \textbf{79.00} & \textbf{79.00} \\ \cline{2-6}
& aw & 89.29          & 90.00 & 89.00 & 89.00 \\ \cline{2-6}
& ay & \textbf{98.21} & \textbf{98.00} & \textbf{98.00} & \textbf{98.00} \\ \hline
\multicolumn{2}{|c|}{\textbf{Mean}} & 89.64 & 90.00 & 89.60 & 89.60 \\ \hline

\end{tabular}
\end{table}

\subsection{RESULTS FOR BCI COMPETITION III IVA DATASET}
Table \ref{tab1} summarizes the accuracy and additional performance metrics of the BCI competition III IVA Dataset for five subjects using different classification algorithms.
For subjects 'aa' and 'aw', MLP achieves maximum classification accuracy of 91.07\% and 96.43\%, respectively. For the subject 'al', SVM achieves a maximum accuracy of 97.86\%. For subjects 'av' and 'ay', XGBoost achieves 78.57\% and 98.21\% accuracy, which is the highest among all other classifiers. At least 0.77\%, 2.56\%, and 1.13\% of the time, the SVM classifier surpasses the MLP, RF, and XGBoost classifiers as far as average classification accuracy is concerned.  Table \ref{tab1} further showed that the SVM classifier reaches its maximum performance for most cases, including precision, recall, and F1 score.  The MLP classifier outperforms the SVM, RF, and XGBoost classifiers with mean precisions of 0.20\%, 1.4\%, and 0.4\%, respectively.  Additionally, the MLP classifier has a mean recall greater than 0.20\%, 1.40\%, and 0.20\%, respectively, of SVM, RF, and XGBoost classifiers. The average F1 score of the MLP classifier also outperforms the SVM, RF, and XGBoost classifiers at least by 0.20\%, 2.80\%, and 0.20\%, respectively.
Experimental results using the   BCI competition III IVA Dataset confirm the superiority of the proposed method over conventional and state-of-the-art approaches as demonstrated in Table \ref{Comparison_with_state_of_the_art_dataset1}. Notably, the proposed method achieved the highest mean classification accuracy of 90.77\%, outperforming several benchmark methods such as stdWC (88.80\%), CSCC (88.62\%), and R-MDRM (87.21\%). The use of embedded feature selection via Random Forest helped reduce feature dimensionality and redundancy, further improving classification performance and computational efficiency. Among the four classifiers, SVM consistently delivered the highest accuracy, precision, recall, and F1 score in most subjects. This suggests that SVM, with its ability to handle higher dimensional feature spaces and generalize well with fewer training samples, is highly suitable for EEG-based MI classification tasks.
The proposed method also demonstrated robustness across individual subjects, achieving consistently high performance even for subjects with previously challenging EEG patterns. The inclusion of spatial filtering through CSP, fuzzy clustering through FCM, and geometric transformations via TSM contributed to a rich and discriminative feature set. Moreover, the focus on anatomically informed channel selection helped minimize the number of EEG channels required, enhancing the method's practicality for real-world BCI applications.
Furthermore, comparative analysis with other methods reveals a significant improvement in accuracy (ranging from 1.97\% to 4.9\%) over existing approaches. This highlights the advantage of integrating multi-domain features with brain region specificity, which is often overlooked in traditional BCI systems. Figures \ref{conventional_vs_proposed_dataset1} illustrate the classification performances of MI-BCI utilizing both the conventional method and the proposed method for BCI competition III-IVa.

\begin{table}[t!]
\centering
\caption{Evaluation of accuracy performance relative to leading studies on BCI competition III IVA Dataset.  The optimal outcome is highlighted in boldface.}
\label{Comparison_with_state_of_the_art_dataset1} \renewcommand{\arraystretch}{0.6}
\resizebox{0.50\textwidth}{!}{%
\begin{tabular}{|c|c|c|c|c|c|c|c|}
\hline
\textbf{Methods} & \textbf{Year} & \multicolumn{5}{c|}{\textbf{Subjects Accuracy (\%)}} & \textbf{Mean ± STD} \\ \hline
                 &               & aa      & al      & av       & aw       & ay       &                     \\ \hline

CSCC \cite{b38}        & 2020          & \textbf{89.29} & 98.21 & 73.47    & 92.86    & 89.29    & 88.62 ± 8.25        \\ \hline
CSP-R-MF \cite{b16}     & 2021          & 82.14   & 96.42   & 72.14    & 84.38    & 94.28    & 85.87 ± 9.83        \\ \hline
LRFCSP \cite{b39}       & 2019          & 83.93   & 96.42   & 74.49    & 88.84    & 89.29    & 86.59 ± 8.10        \\ \hline
CCS-RCSP \cite{b40}     & 2019          & 83.03   & 96.42   & 70.91    & 92.41    & 92.46    & 87.05 ± 10.28       \\ \hline
SR-MDRM \cite{b23}  & 2019          & 79.46   & 100.00  & 73.46    & 89.29    & 88.49    & 86.13 ± 9.09        \\ \hline
R-MDRM\cite{b25} & 2019          & 81.25   & \textbf{100.00} & 76.53 & 87.05 & 91.26    & 87.21 ± 8.12        \\ \hline
FCCR\cite{b41} & 2018          & 78.57   & 98.21   & 72.45    & 87.05    & 93.25    & 85.90 ± 10.51       \\ \hline
BECSP\cite{b14} & 2022          & 77.68   & 100.00  & 73.98    & 84.82    & 88.10    & 84.91 ± 9.05        \\ \hline
stdWC\cite{b37} & 2023          & 84.50   & 98.10   & 72.80    & 95.10    & 92.50    & 88.80 ± 9.10        \\ \hline
\textbf{Proposed Method}  & 2025          & 87.14   & 97.86   & \textbf{77.50} & \textbf{96.07} & \textbf{95.36} & \textbf{90.77 ± 7.60} \\ \hline
\end{tabular}%
}
\end{table}

\subsection{RESULTS FOR BCI COMPETITION IV I DATASET}
The performance evaluation of BCI competition IV-1 Dataset, as shown in Table \ref{tab2} across multiple classifiers demonstrates that the Support Vector Machine (SVM) achieved the best overall results, with the highest mean accuracy of 84.5\% and balanced precision 82.0\%, recall 81.5\%, and F1-score 81.6\%. This indicates that SVM provided consistent and reliable classification across subjects. The MLP followed closely with a mean accuracy of 82.1\% and nearly identical precision, recall, and F1-scores, highlighting its stability, although performance declined notably for subject f. Random Forest (RF) achieved a mean accuracy of 77.9\% with reasonably high precision of 79.0\%, but lower recall of 77.9\% and F1-score of 77.7\%, suggesting weaker detection capability and reduced balance between metrics. XGBoost performed the worst, with a mean accuracy of 76.4\% and the lowest precision of 76.9\%, recall of 76.3\%, and F1-score of 76.1\%, indicating limited generalization capability. At the subject level, classification was consistently easiest for subject e 95\% across models, while subject 'f' proved the most challenging, with significantly lower accuracy across all classifiers. Overall, the results suggest that SVM is the most effective classifier for this dataset, followed closely by MLP, whereas RF and XGBoost lag behind in performance.

\begin{table}[t!]
\caption{The evaluation of the proposed approach regarding classification accuracy, precision, recall, and F1 score utilizing MLP, SVM, RF, and XGBoost classifiers using BCI competition IV-1 Dataset. The most significant outcome is highlighted in boldface.}
\label{tab2} \renewcommand{\arraystretch}{0.7}
\centering
\begin{tabular}{|c|c|c|c|c|c|}
\hline
\textbf{Classifier} & \textbf{Subjects} & \textbf{Accuracy} & \textbf{Precision} & \textbf{Recall} & \textbf{F1 Score} \\ \hline

\multirow{7}{*}{MLP} 
& a & 82.50 & \textbf{83.00} & 82.00 & 83.00 \\ \cline{2-6}
& b & 72.50 & 73.00 & 72.00 & 72.00 \\ \cline{2-6}
& c & 75.00 & \textbf{75.00} & \textbf{75.00} & \textbf{74.00} \\ \cline{2-6}
& d & \textbf{95.00} & \textbf{95.00} & \textbf{95.00} & \textbf{95.00} \\ \cline{2-6}
& e & \textbf{95.00} & \textbf{95.00} & \textbf{95.00} & \textbf{95.00} \\ \cline{2-6}
& f & 67.50 & 68.00 & 68.00 & 68.00 \\ \cline{2-6}
& g & \textbf{87.50} & \textbf{88.00} & \textbf{88.00} & \textbf{88.00} \\ \hline
\multicolumn{2}{|c|}{\textbf{Mean}} & 82.14 & \textbf{82.43} & \textbf{82.14} & \textbf{82.14} \\ \hline

\multirow{7}{*}{SVM} 
& a & \textbf{82.50} & 86.00 & \textbf{85.00} & \textbf{86.00} \\ \cline{2-6}
& b & 78.50 & 75.00 & 75.00 & 75.00 \\ \cline{2-6}
& c & 77.00 & 71.00 & 70.00 & 70.00 \\ \cline{2-6}
& d & 90.00 & 90.00 & 90.00 & 90.00 \\ \cline{2-6}
& e & 92.46 & \textbf{95.00} & \textbf{95.00} & \textbf{95.00} \\ \cline{2-6}
& f & \textbf{86.50} & \textbf{81.00} & \textbf{80.00} & \textbf{80.00} \\ \cline{2-6}
& g & 84.50 & 76.00 & 75.00 & 75.00 \\ \hline
\multicolumn{2}{|c|}{\textbf{Mean}} & \textbf{84.50} & 82.00 & 81.45 & 81.57 \\ \hline

\multirow{7}{*}{RF} 
& a & 75.00 & 78.00 & 75.00 & 75.00 \\ \cline{2-6}
& b & \textbf{80.00} & \textbf{80.00} & \textbf{80.00} & \textbf{80.00} \\ \cline{2-6}
& c & 72.50 & 73.00 & 72.00 & 73.00 \\ \cline{2-6}
& d & 87.50 & 88.00 & 88.00 & 88.00 \\ \cline{2-6}
& e & 90.00 & 90.00 & 90.00 & 90.00 \\ \cline{2-6}
& f & 60.00 & 64.00 & 60.00 & 58.00 \\ \cline{2-6}
& g & 80.00 & 80.00 & 80.00 & 80.00 \\ \hline
\multicolumn{2}{|c|}{\textbf{Mean}} & 77.86 & 79.00 & 77.86 & 77.71 \\ \hline

\multirow{7}{*}{XGBoost} 
& a & 70.00 & 71.00 & 70.00 & 70.00 \\ \cline{2-6}
& b & 80.00 & \textbf{80.00} & \textbf{80.00} & \textbf{80.00} \\ \cline{2-6}
& c & 72.50 & 73.00 & 72.00 & 71.00 \\ \cline{2-6}
& d & 85.00 & 86.00 & 85.00 & 85.00 \\ \cline{2-6}
& e & \textbf{95.00} & \textbf{95.00} & \textbf{95.00} & \textbf{95.00} \\ \cline{2-6}
& f & 57.50 & 58.00 & 57.00 & 57.00 \\ \cline{2-6}
& g & 75.00 & 75.00 & 75.00 & 75.00 \\ \hline
\multicolumn{2}{|c|}{\textbf{Mean}} & 76.43 & 76.86 & 76.29 & 76.14 \\ \hline

\end{tabular}
\end{table}
\begin{table*}[t!]
\caption{A comparison of performance in terms of accuracy with state-of-the-art studies on the BCI competition IV-1 Dataset, with the optimal outcome highlighted in boldface.}
\label{Comparison_with_state_of_the_art_dataset2}\renewcommand{\arraystretch}{0.7}
\centering
\resizebox{\textwidth}{!}{%
\begin{tabular}{|c|c|c|c|c|c|c|c|c|c|}
\hline
 &  & \multicolumn{7}{c|}{ \textbf{Subjects Accuracy}} &  \\ \cline{3-9}
\multirow{-2}{*}{\textbf{Methods}} & \multirow{-2}{*}{\textbf{Year}} & a & b & c & d & e & f & g & \multirow{-2}{*}\textbf{{Mean ± STD}} \\ \hline
TFPO-CSP {[}44{]} & 2018 & \textbf{87.50} & 55.00 & 67.50 & 85.00 & 87.50 & 85.00 & 92.50 & 80.00±6.18 \\ \hline
R-AdaBoost {[}45{]} & 2019 & 83.50 & 62.00 & - &-  & - & 76.59 & 91.50 & 78.40±10.84 \\ \hline
CCS-RCSP {[}46{]} & 2019 & 86.50 & 53.50 &  - & - & -  & 66.50 & \textbf{93.50} & 75.00±15.88 \\ \hline
BCS-CSP {[}47{]} & 2020 & 78.50 & 77.50 &  - & - & -   & \textbf{92.00} & 87.00 & 83.80±6.03 \\ \hline
CSRI {[}48{]} & 2023 & 72.80 & 66.18 & \textbf{83.72} & \textbf{96.10} & 83.50 & 76.33 & 87.33 & 80.85 ± 9.18 \\ \hline
Entropy-based Channel Selection {[}49{]} & 2024 & 80.00 & 67.00 & 80.00 & 91.00 & 90.00 & 91.00 & 80.00 & 82.79 ± 8.73 \\ \hline
\textbf{Proposed Method} &  & 82.50 & \textbf{78.50} & 77.00 & 90.00 & \textbf{92.46} & 86.50 & 84.50 & \textbf{84.50}±7.60 \\ \hline
\end{tabular}%
}
\end{table*}


The comparative analysis using  BCI competition  IV-I Dataset is demonstrated in Table \ref{Comparison_with_state_of_the_art_dataset2}, which shows that the proposed method achieves a balanced and competitive performance across all subjects, with an average accuracy of 84.50\%. This outperforms most existing methods, such as TFPO-CSP (80.00\%), R-AdaBoost (78.40\%), CCS-RCSP (75.00\%), and CSRI (80.85\%). Only Entropy-based Channel Selection (82.79\%) and BCS-CSP (83.80\%) come close, but the proposed approach still demonstrates higher overall stability and accuracy.
Looking at individual subject performance, the proposed method performs exceptionally well for subjects 'd' (90.00\%), 'e' (92.46\%), and 'f' (86.50\%), while maintaining consistent accuracy for the other subjects without large fluctuations. In contrast, earlier methods such as R-AdaBoost and CCS-RCSP show greater variability, with weaker results in some subjects (e.g., subject 'b' with 62.00\% and 53.50\%, respectively). Figures \ref{conventional_vs_proposed_dataset2} illustrate the classification performances of MI-BCI utilizing both the conventional method and the proposed method for BCI Competition IV-I.

\section{DISCUSSION}
\begin{figure}[t!]
\centering
\includegraphics[width=0.38\textwidth]{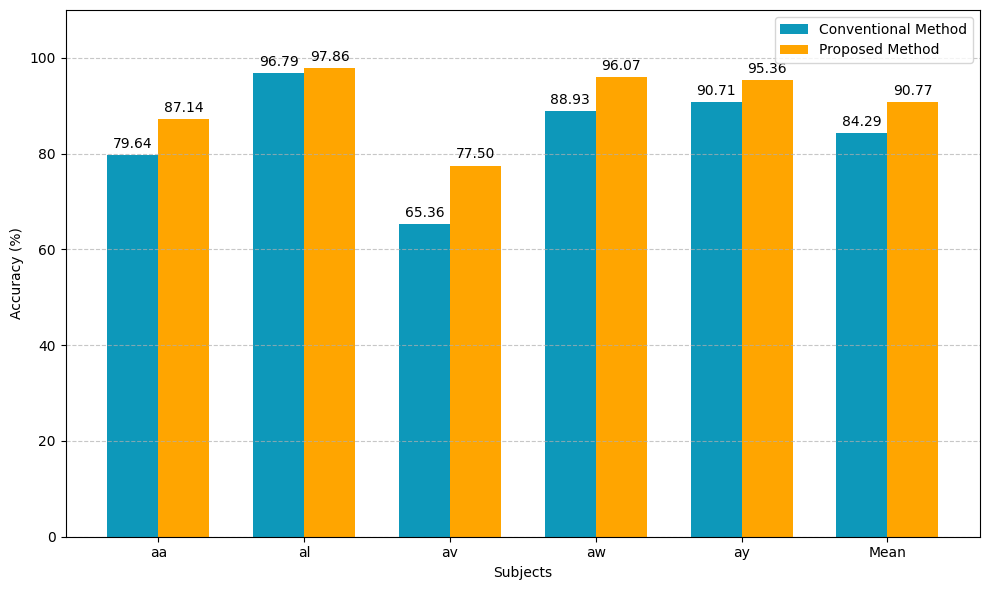}
\caption{Comparison of performance (classification accuracy) between the proposed method and the conventional MI classification method using the BCI competition III IVA Dataset.
}
\label{conventional_vs_proposed_dataset1}
\end{figure}

\begin{figure}[t!]
\centering
\includegraphics[width=0.38\textwidth]{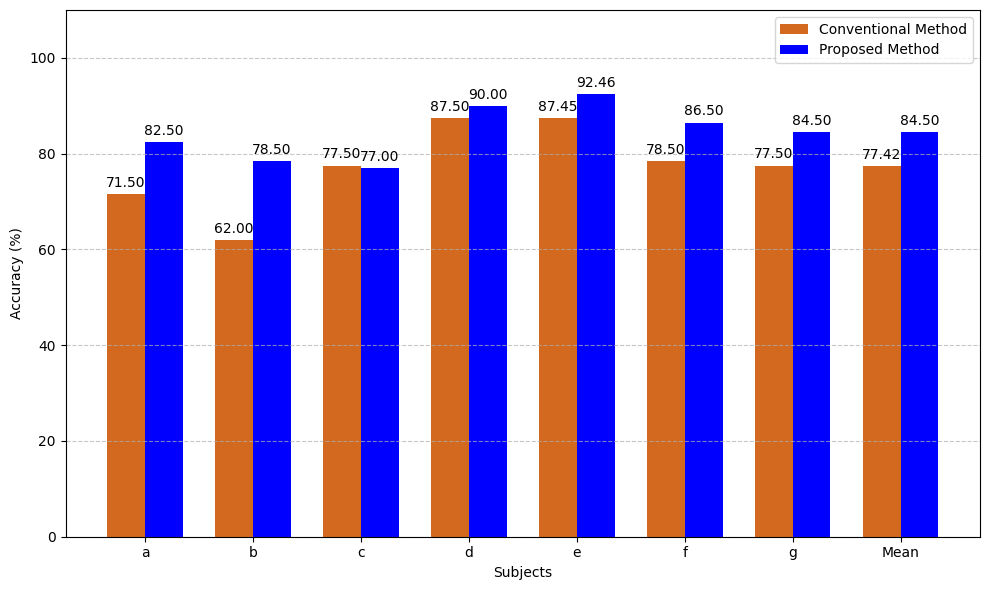}
\caption{Comparison of performance (classification accuracy) between the proposed method and the conventional MI classification method using the BCI competition IV-1 Dataset.
}
\label{conventional_vs_proposed_dataset2}
\end{figure}

Figures \ref{conventional_vs_proposed_dataset1} and \ref{conventional_vs_proposed_dataset2} compare the classification performance of MI-BCI using both the conventional method and the proposed approach for the BCI Competition III-IVa and BCI Competition IV-I datasets. The conventional method utilizes all channels, but its performance consistently lags behind the proposed method. The analysis reveals that the multi-domain feature fusion approach enhances performance both individually and on average across all subjects. The proposed method, using SVM, achieves a classification accuracy that surpasses the conventional approach by 6.48
The application of cross-validation leads to some variation in the results across individual repetitions, with the final accuracy for each subject being the average of all repetitions. The results of this study show that the proposed brain region-based multi-domain feature integration technique enables effective motor imagery (MI) classification using EEG signals. By clustering EEG channels based on their anatomical relevance to motor function and extracting spatial, spectral, and geometric features, the method captures diverse and complementary information essential for MI decoding. These findings underscore that the proposed method not only improves mean classification accuracy but also provides more reliable performance across different subjects, making it a promising solution for real-time BCI applications, especially in assistive technologies and neuro rehabilitation.

\section{CONCLUSION}
This study introduced a brain region-based multi-domain feature fusion framework for EEG-based motor imagery (MI) classification. By grouping EEG channels according to their functional brain regions and applying three complementary feature extraction techniques—Common Spatial Pattern (CSP), Fuzzy C-Means clustering, and Tangent Space Mapping (TSM)—the proposed method effectively captures spatial, clustering, and non-linear characteristics of EEG signals. The integration of these features into a high-dimensional vector, followed by SVM-based classification, significantly enhanced recognition performance. Experimental validation on benchmark datasets (BCI Competition III dataset I and BCI Competition IV dataset IVA) demonstrated consistent improvements over conventional approaches, achieving average classification accuracies of 84.50\% and 90.77\%, respectively. The framework also reduced channel dependency and computational cost, underscoring its suitability for real-time BCI applications. Overall, the results confirm that combining brain region-specific channel selection with multi-domain feature fusion strengthens both the robustness and interpretability of MI-based BCI systems. Future work will extend this framework to multiclass MI scenarios and investigate its integration with deep learning models for further performance gains.


\begingroup
\small   
\bibliographystyle{unsrt}
\bibliography{Reference}
\endgroup

\end{document}